# Implementation of general formal translators

Iosif Iulian Petrila*

Faculty of Automatic Control and Computer Engineering, Gheorghe Asachi Technical University of Iasi, Str. Dimitrie Mangeron, Nr. 27, 700050, Iasi, Romania

The general translator formalism and computing specific implementations are proposed. The implementation of specific elements necessary to process the source and destination information within the translators are presented. Some common directives or instructions, such as classes and procedures, were unified and generalized in order to allow general translations implementations. In order to cover general cases, two levels of processing are required, related to the source and destination information appropriate transformations, with the related control and processing instructions. The proposed general translator elements are useful for processing natural or artificial information described through any types of languages or systems.

## Introduction

One of our basic needs or of any other distinct social natural beings is to communicate in order to exchange and correlate information, the similar needs also exist within artificial systems under all the aspects in which the information is represented, transported, modeled, processed, coded, encrypted etc [1-4]. However, one of the most important challenges is to achieve the translation of information from one mode of representation to another, in a general way, which is still an open question in many fields as: linguistics, information science, computer science, computer engineering etc.

The way in which the information can be accessed and used is through a language, which in a general sense is nothing else than a set of rules (conventions, operations, laws, phenomena etc.) and some related symbols (numerical, graphical, audio, objects, particles etc.) used to represent its components. From this perspective, the languages are the instruments that facilitate the description and the management of information related to any system [5-7].

Different perspectives of the same information can be highlighted by different languages, as well as certain information transformations can also be described through a language. The connections between these descriptions offered by different languages are made by the translators. Nowadays, in order to be effective and relevant, any type of translator must be implemented through a computer system. More than that, a general computerized translator must be based on and also to generalize the best performing translator within computing systems, which is the compiler. The compiler, as the fundamental translator within computing systems, must itself first to be abstracted and generalized in order to allow the augmentation and flexibilisation of the usual programming languages and to facilitate the approach and management of any languages and translations categories, specific to computing systems (such as compilation/decompilation, encryption/decryption, compression/decompression, obfuscation/lisibilization etc.) or to natural systems [8-17].

A versatile translation framework must allow a unitary management of existing languages (natural or artificial) but also must be able to cover the present and the future implementation challenges related to all categories of systems such as: quantum, neural, smart, social, linguistic, algorithmic, chemical etc [18-26].

## General Translators

A general translator must able to convert any sources of information descriptive through a language in any other type of description or representation. Effectively, directly or indirectly/implicitly, the translation process must involve at least two steps of processing, one focused on source and other on destination. For example, in the case of the web client-server information transfer, at least one level of processing exists in the server (source) area (back-end) and at least one in the client (destination) area (front-end). In the case of compilation, one level is involved at the input processing stage (high-level user facile descriptions) and another (as compilation directive, make file operations etc.) in the output formatting stage (low-level machine format representation). The same two levels must be also involved in the case of natural languages in the translation from one language to another or even internally in the formation of representations or perceptions of some information. From this perspective, a general translator must provide hierarchically-organized environments and tools in order to analyze and manage various source information described through arbitrary languages and transformed in a different informational structure or language description. The computer implementable requirement of such translators must involve and to generalize the concept of compiler, this being the fundamental translator used in computing systems. From this perspective, a formal general translator **T** can be defined by

$$T = \langle S, D, I, \sigma, \delta \rangle$$

with **S** – source language alphabet, **D** – destination language alphabet, **I** internal or intermediate alphabet (**I** ⊆ **S**, **I** ⊆ **D**), σ – source language rules, δ – destination language rules. Some internal transformations or rules ι (which would operate with elements from **I**) can also be included in the definition (corresponding to middle end processing) but these are in fact manageable through σ and/or δ. The **T** translator can also be detailed with the elements specific to Turing machines formalism [3] with (tape, states etc.) as a two interconnected Turing machines. From implementation perspective, the solution of a reliable general translator is given by highlighting the two distinct levels of processing corresponding to at least two languages associated with the source and destination information. Also at each level is necessary two types of explicit control instructions (decision sublevels), similar to some explicit or implicit control instructions used in some common compilers one at high-level (for preprocessing, syntax operations etc.) and one at low-level (for compile/assembly/linking/optimization/formatting/postprocessing operations). A diagram with the general stages and operations specific to the general translations can be seen in Figure 1.

* **E-mail address:** IosifIulianPetrila@gmail.com

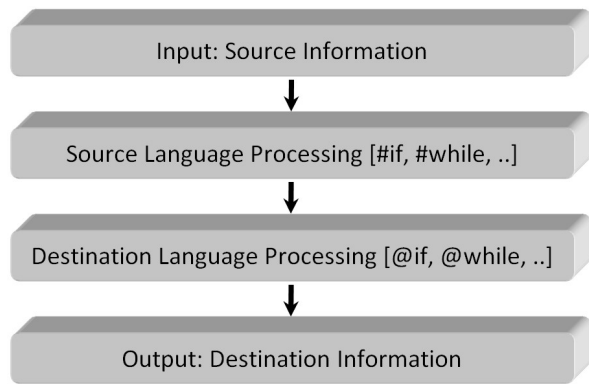

Figure 1. Translation of source to destination

Each processing level works as the usual language processing with typical directives, instructions etc. At source level is performed a decomposition of the source information described by the source language (which cover σ rules) where the most relevant keys are prefixed by **#** (`#if`, `#elif`, `#else`, `#endif`, `#while`, `#endw`, `#repeat`, `#until`, `#break`, `#error`, `#print` etc.). At destination level is performed a composition of the information according to the related destination language instructions (incorporated in δ rules), where the most relevant keys are prefixed by **@** (`@if`, `@elif`, `@else`, `@endif`, `@while`, `@endw`, `@repeat`, `@until`, `@break`, `@error`, `@print` etc.). At both levels, can be used relational and equivalence expressions/operations (`<` – less than, `>` – greater than, `<=` – less than or equal to, `>=` – greater than or equal to, `=` – equal to, `!=` or `<>` – not equal to), logical expressions/operations (`||` – or, `&&` – and, `!` – not), bitwise expressions/operations (`&` – and, `|` – or, `^` – xor, `~` – complement, `<<` – left shift, `>>` – right shift) and arithmetic expressions/operations (`+` – addition, `-` – subtraction, `*` – multiplication, `/` – division, `%` – modulo/remainder) etc.

**Source Language Processing**

The source information described through the source language is processed first within the first level of processing (depending on the relative human factor perspective in relation to the source or destination, in some translators this level of processing it can be found under different names such as high-level, user-level, machine-level, front-end, back-end etc.) designated to implement and manage the lexical and syntactic rules of demand source language and also to convert the information in a unitary internal form for processing and conversion to the destination information. For example, in the case of compilers, source language processing corresponds to the high-level/user-level/front-end processing, in the case of decompilers corresponds to the low-level/machine-level processing, while in the case of the server-browser system translator it corresponds to the server-level/back-end processing and so on. At this level, the information must be decomposed into components through general but also language specific directives and instructions.

The decision instructions are considerate as line prefixed directive, similar to (macro) instructions used in languages such as C/C++ [8]. For instance, the typical `if-else` instruction can be used as:

```
#if condtition1
...
#elif condtition2
...
#else
...
#endif
```

The `condtition1`, `condtition2` have a logical/conditional structure with the usual operators and meanings. The simple `while` loops can be used as:

```
#while condtition
```

```
...
[#break [levels]]
...
#endw
```

where `#break` directive may be optionally followed by the `levels` (to breaks) as a number of loops blocks to break (0 - do nothing or can be assumed as continue, 1 - break current block, 2 - break also parent block etc.). In some situations is useful the `repeat`-`until` loop instruction eventually times limited, which can be used as:

```
#repeat [times]
...
[#break [levels]]
...
#until [condition]
```

where `times` is the maxim repeated loops if `condition` remains false or is absent.

If something goes wrong during the source information processing, the translation may be stopped by using

```
#error message
```

with a `message` to be displayed. Also, at this level of processing, the non error messages can be displayed with a similar instruction:

```
#print message
```

Identifying the specific syntactic components of a language, with the related semantics, is one of the most complex operations within translators. These operations are carried out through the general `class` implemented as a hyperclass or hyperfunction which unites in a single concept the syntax parser, (class/object) structures, procedures, classes, hygienic macro functions, syntactic closures etc. and allowing their definition but also the management of their invocations. This generalized `class` is defined by

```
class name [fargs [.. vargs]]
{
   ...
}[name1 [... , nameN]]
```

with `fargs` fixed arguments, `vargs` variable/repetitive arguments (with `..` as separator operator of fixed arguments from variable/repetitive ones) and `name1` .. `nameN` some alias names of the current defined `name` `class`. Each argument can be delimited by one or more symbol operators (`+`, `*`, `|` .. even comma `,` etc.), in this way, this `class` generalizes the concept of lambda calculus by using multiple arguments with any type of separators to a syntactic general lambda calculus in which the call arguments can be separated by any type of operator, allowing the specific management of a general syntactic parser. The `class` instruction can handle any operation declared by using prefix, postfix or infix notation. A `class` can be multiple rewritten/overriding/redefined later, with the same or different types of arguments and separators, without warning because the global context may be different and useful to be handled in each situation etc. For instance, if we define first

```
class Sum x + y {S1(x, y)}
```

and later

```
class Sum x + y {S2(x, y)}
```

the last defined `class` is used first when `Sum` `class` is called (the result will be `S2(x, y)`) if the specific global context is the same. To keep simple and natural the implementation of various operations, a `class` can be also overdefine / overload / overwrite. Also, a `class` can be called or invoked as a usual procedure or function. When a `class` with a given name is calling, is verify first if the calling arguments are match from last to first defined `class`. In contrast to the usual approaches of functional languages (and not only) in which the verification of corresponding arguments in multiple definitions is made from the first definition to the last, the present classes will be verified in the reverse order of their definition, from the last definition to the first. The main argument is that the novelty appears later and must be present also after the previous definitions. In this respect, the more general classes should be placed after the less general ones etc. In this way, the problem of upgrading of a library is solved by including first the usual code and later adding new characteristics etc. Also, the operations priorities definition and management are simply solved by this order definition rule, in this respect, a syntax `S` `class` with different arguments which must be defined in an order that is correlated with the priority of the involved operators

```
class S X1{...}
...
class S Xn{...}
```

where `X1` .. `Xn` are argument-expression (may contain symbol operators). When is used `S` `class`

```
S Expression
```

it is verify each `S class` from last to first defined and is used only the `class` with argument which `Expression` is match. An example of an algebraic `class` which parses simply addition and multiplication expression can be defined as:

```
class E x {... x ...}
class E x * y {... E(x) * E(y) ...}
class E x + y {... E(x) + E(y) ...}
```

In this case, when `E class` is used with an expression which contain both operations '`+`' and '`*`', as in

```
E a * b + c
```

for instance, the additive operation is identified and processed first through the last definition of the `class E` (`x` is `a * b` and `y` is `c` according to last `E class`) and second the multiplicative operation, according to operations relative priority/precedence which imposed the order/succession of `E` classes definitions.

In some situations is need to parse multiple arguments in a specific order. For instance, simple integer addition operation of multiply arguments is useful to be implemented by adding each argument into an accumulator register etc. For the operations reducible to binary operation, the associative property can be implemented as left-associative operation `x * y * z = (x * y) * z` or as right-associative operation `x * y * z = x * (y * z)`. An example of implementation of a left-associative operation is:

```
class E x {... x ...}
class E x * y {... E(x)*E(y) ...}
class E x * y * z
{... E(x*y)*E(z) ...}
```

A right-associative operation is obtained by replacing the last class with

```
class E x      *     y     *     z
{... E(x)*E(y*z) ...}
```

The `class` instruction can be used to define various types of information like: simple or structured data, procedures, objects etc. For instance, one can define simple `Long` data as:

```
class Long L
{
   L dd ?
}
```

where, in this case, `Long` it was defined as double-word through `dd`. A structured data, like a 2D point for instance, can be defined as:

```
class Point P
{
   Long P.x
   Long P.y
}
```

where, inside of `Point class` body, each field it was defined as `Long`. One may mention that `P` argument (which will in fact be the next variable names defined by `Point`) is substituted inside a body `class` on all operand in doted (in this case) or other operation. In the same way can be used `class` instruction to define any data type according with demand programming or natural language. For example, object classes used in object-oriented languages (with encapsulation, inheritance, polymorphism and abstraction features), can be easily defined as:

```
class Parent P
{
   ...
   Long P.a
   ...
}

class Child C
{
   Parent C
   ...
   Point C.p
   ...
}

Child V
```

where, for `V Child class` variable, the characteristic of its parent `Parent` are inherited and accessible as `V.a` etc. One may observe that the inheritances (simple or even multiple) are specified in the body of the `class` where there is more freedom and flexibility to operate with them and not outside as in the case of the usual programming languages.

To cover even the most complex information (such as template structures etc.) it allowed defining a `class` inside to another `class` as:

```
class c a
{
   ...
   class a ... {...}
   ...
```

```
  class b ... {...}
  ...
}
```

where `a class` is defined according with called argument of `c class` and `b class` may contain only inside references to the calling argument `a`. For instance, if we want to implement typical C `#define` preprocessor directive macro definition, the following definitions can be used

```
class #define a b {a := b}

class #define a(x) b {class a(x){b}}
```

where, first are solved functional replacements (if any) and second the symbol substitution. In the case of functional replacement `a(x)`, `x` can contain one argument or more and no additional definition is required. In a similar way can be implemented the specific elements of lambda calculus or any other processing element.

Along with the syntactic analysis facility, the `class` instruction generalizes and incorporates the usual programming languages macro functions, structures, classes, procedures with nested and closure characteristics etc. With this flexible `class` instruction, near to control instructions, it can be handled the syntax of even the most complex programming or natural language or different type of data or information. In relation to the way in which the analysis of source language syntax is performed, the present approach is: implicit, dynamic, contextual and implementable. At this level, after splitting the source information into components, a series of operations may be necessary to identify information semantics, these classes can be also used for this purpose, regardless of the type of method used in this sense as neural etc.

Because at source level processing it converts the source information into internal abstract representation or partial destination information, the next destination level operations will allow the organization of information in accordance with the specific rules of the destination language similar to the operations of usual compilers in the sage of output formatting operations (assembly, linking, optimization etc.).

**Destination Language Processing**

At destination level processing/stage, the information/code is transformed into the desired output format according to destination language and information system characteristics, similar to processor and operating system type and generally corresponds to the middle-end and back-end usual compiler operations. The explicit existence of control instructions within this level is essential because the formatting of the output cannot be done by simple operations, for these operations the usual compilers are used some extra languages or directives in a implicitly and non-transparently way, such instructions are used incoherently within the usual compilers, often through some scripts and directives less correlated with the compiled language etc. In addition, a series of optimization operations are required, which require the use of decision and control instructions. This second category of transformations is often implicitly included in the linking routines in the case of programming languages information processing (such as C/C++ for which both the source and destination languages are clear etc.). In the general case, for the translation of information described from any language to any other language, these translations must be explicitly described by instructions that facilitate optimization operations specific to the target language. In this respect, similar to high-level processing, with equivalent definitions and meanings, the destination language information processing (low-level languages in the case of compiling information described by the programming languages) must use similar control instructions: `@if-@elif-@else-@endif`, `@while-@break-@endw`, `@repeat-@break-@until` etc. The control instructions are also very useful for destination information processes (to perform code optimization, managing multiple processing steps etc.) but in an explicit way. For instance, if some procedures are not used (called) during translation/compilation, then one may add a conditional translation/compilation like:

```
@if [Pn]
  ...
  Pn:
  ...
@endif
```

which include implementation of the procedure only if the entry point in procedure `Pn` (calling name or label) is used elsewhere, useful to eliminate the code/body of unused/uncalled procedures etc. In some languages, these types of instructions can be replaced by operations external to the language performed by the compilers, or they can even be replaced by some preprocessing instructions [8].

In case of possible errors that may occur during this level processing stage, the execution of the translator may be stopped by using:

```
@error message
```

with a `message` to be displayed. The non error messages can be displayed also at this level of processing with

```
@print message
```

One may mention that, in the case of using both `#print` and `@print` instructions, the next two messages will be displayed in reverse order because at the first processing level/stage `#print` will be managed and `@print` at the next level/stage.

```
@print "Output Processing ..."
#print "Input Processing ..."
```

An important process in relation to this stage is to write on to destination various types of information and values on different bytes size representation:

```
dX v1 [[,] v2 ...]
```

as the usual low-level specific data instructions, where `X` can be `b` for byte, `w` for word (two bytes), `d` for double word (four bytes), `p` for three words (six bytes), `q` for quad words (eight bytes) etc. If we want to reserve for instance only upper case chars one may use

```
I = 'A'
@while I <= 'Z'
  db I
  I = I + 1
@endw
```

where in the output will be write in this case `ABCD..Z`. In the same way it can be write in the output repeated or reserved multiple data units through

```
rX N [[,] V]
```

where `X` has the same meaning as in the above definition, `N` is the number of allocation units and `V` (if present) is the values to fill each unit.

The connection between source information or source language elements and the destination correspondents is performed through combining source directives or instructions with destination ones. For example, in the simplest case of translation from assembly language to machine language, "No Operation Performed" - `nop` instruction can have the following simply implementation for Intel (x86) system

```
class nop {db 0x90}
```

or one of the following implementations for the ARM/RISCV systems

```
class nop {dd 0}//andeq r0, r0, r0
class nop {dd 0xE1A00000}//mov r0, r0
class nop {dd 0x13}//addi x0, x0, 0
```

In the same way can be implemented and managed the instructions related to any languages for translating any types of language elements such as: literals, operations, expressions, phrases etc. The present implementation approaches can be used to implement any type of translator and also can provide clues for completing some source or destination languages with elements that allow them to be flexible, versatile, optimally implementable etc [8].

**Conclusions**

The general formalism related to the description and implementation of general translators has been proposed. The formal descriptions of the translators were correlated with computer implementative elements by generalizing some computational concepts and systems in order to facilitate the translation of any source information, described according to the source language rules, into destination information structured according to other language rules.

The most difficult problem related to the translation of some arbitrary languages through general parsing rules by splitting source information into its basic components was solved by generalizing and joining the concepts of function/procedure and object/class into a hyperclass with the role of hyperfunction that accepts arguments with any type of separator, separators that can be source language operators/delimiters/punctuators, in this way the syntactic components source splitting issue being resolved.

The proposed general translator system is based on compiler generalization through incorporating and detailing source and destination language operations by using two levels of processing related to the source and destination information specific transformations, with the related control and processing instructions. The two levels of processing are proposed in order to cover the general translations cases, but for certain translation, it is possible that some instructions from a certain level to not be explicit necessary as is in the case of some usual compilers where they are used implicitly or as auxiliary or external operations.

The presented characteristics of the general translators allow procession of information within the computing systems of any type of natural language or programming/script language facilitating implementations through any operating systems or processors. Some of the presented elements may contribute to optimization of some usual translators such as compilers and related languages as well as for more relevant translations of natural languages descriptions.

## References

[1] N. Chomsky, *Three models for the description of language*, IRE Transactions on Information Theory 2, 113-124 (1956).

[2] C. E. Shannon, *A Mathematical Theory of Communication*, Bell System Technical Journal 27, 379-656 (1948).

[3] A. M. Turing, *On computable numbers, with an application to the Entscheidungsproblem*, Proceedings of the London Mathematical Society 42, 230-265 (1936).

[4] J. von Neumann, *First Draft of a Report on the EDVAC (Electronic Discrete Variable Automatic Computer)*, Moore School of Electrical Engineering, University of Pennsylvania (1945).

[5] D. M. Ritchie, *The Development of the C Language*, ACM 28, 201-208 (1993).

[6] T. X. Sun, X. Y. Liu, X. P. Qiu, X. J. Huang, *Paradigm shift in natural language processing*, Machine Intelligence Research 19, 169-183 (2022).

[7] D. Khurana, A. Koli, K. Khatter, S. Singh, *Natural language processing: State of the art, current trends and challenges*, Multimedia Tools and Applications, 1-32 (2022).

[8] I. I. Petrila, *@C – augmented version of C programming language*, arXiv:2212.11245 (2022).

[9] M. Maronese, L. Moro, L. Rocutto, E. Prati, *Quantum compiling*, Quantum Computing Environments, 39-74 (2022).

[10] M. De Coster, D. Shterionov, M. Van Herreweghe, J. Dambre, *Machine Translation from Signed to Spoken Languages: State of the Art and Challenges*, arXiv:2202.03086 (2022).

[11] K. Georgiou, Z. Chamski, A. A. Garcia, D. May, K. Eder, *Lost in translation: Exposing hidden compiler optimization opportunities*, The Computer Journal 65, 718-735 (2022).

[12] B. Haddow, R. Bawden, A. V. M. Barone, J. Helcl, A. Birch, *Survey of low-resource machine translation*, Computational Linguistics 48, 673-732 (2022).

[13] Ö. Özerk, C. Elgezen, A. C. Mert, E. Öztürk, E. Savaş, *Efficient number theoretic transform implementation on GPU for homomorphic encryption*, The Journal of Supercomputing 78, 2840-2872 (2022).

[14] X. Tang, Z. Zhang, W. Xu, M. T. Kandemir, R. Melhem, J. Yang, *Enhancing Address Translations in Throughput Processors via Compression*, Proceedings of the ACM International Conference on Parallel Architectures and Compilation Techniques 20, 191-204 (2020).

[15] F. Tang, D. Huang, F. Wang, Z. Chen, *Universal Signature Translators*, International Journal of Network Security 23, 1058-1064 (2021).

[16] Y. Hao, Q. Li, C. Fan, F. Wang, *Data storage based on DNA*, Small Structures 2, 2000046 (2021).

[17] M. Li, Y. Liu, X. Liu, Q. Sun, X. You, H. Yang, Z. Luan, L. Gan, G. Yang, D. Qian, *The deep learning compiler: A comprehensive survey*, IEEE Transactions on Parallel and Distributed Systems 32, 708-727 (2020).

[18] R. P. Feynman, *Quantum mechanical computers*, Foundations of Physics 16, 507-531 (1986).

[19] R. A. Frost, *Realization of Natural-Language Interfaces Using Lazy Functional Programming*, ACM Computing Surveys 38, 11-es (2006).

[20] J. M. Shine, M. Breakspear, P. T. Bell, K. A. Ehgoetz Martens, R. Shine, O. Koyejo, O. Sporns, R. A. Poldrack, *Human cognition involves the dynamic integration of neural activity and neuromodulatory systems*, Nature Neuroscience 22, 289-296 (2019).

[21] D. Agarwal, Y. Baba, P. Sachdeva, T. Tandon, T. Vetterli, A. Alghunaim, *Accurate and Scalable Matching of Translators to Displaced Persons for Overcoming Language Barriers*, arXiv:2012.02595 (2020).

[22] W. Merrill, *Formal language theory meets modern NLP*, arXiv:2102.10094 (2021).

[23] P. Lu, R. Gong, S. Jiang, L. Qiu, S. Huang, X. Liang, S. C. Zhu, *Inter-gps: Interpretable geometry problem solving with formal language and symbolic reasoning*, arXiv:2105.04165 (2021).

[24] Y. Hao, D. Angluin, R. Frank, *Formal language recognition by hard attention transformers: Perspectives from circuit complexity*, Transactions of the Association for Computational Linguistics 10, 800-810 (2022).

[25] S. Longo, *Generative grammars for branched molecular structures*, Chemical Physics Letters 809, 140151 (2022).

[26] T. Æ. Mogensen, *Programming Language Design and Implementation*, Springer Nature, 2022.